\documentclass{article}
\usepackage{microtype}
\usepackage{graphicx}
\usepackage[hypcap=true]{caption}
\usepackage{subcaption}
\usepackage{wrapfig,enumitem,xspace}
\usepackage{booktabs} % for professional tables

\usepackage{tabularx}
\PassOptionsToPackage{numbers}{natbib}
\usepackage{natbib}
\usepackage{float}
\usepackage{hyperref}

\newcommand{\ours}{SAPG\xspace}

\usepackage[accepted]{icml2024}

\usepackage{amsmath}
\usepackage{amssymb}
\usepackage{mathtools}
\usepackage{amsthm}
\usepackage[capitalize,noabbrev]{cleveref}

\theoremstyle{plain}

\theoremstyle{definition}

\theoremstyle{remark}

\usepackage[textsize=tiny]{todonotes}

\icmltitlerunning{SAPG: Split and Aggregate Policy Gradients}

\begin{document}

% Distributed collaborative optimization (DCO)
% Parallel agent coordination (PAC)

\twocolumn[{%
\renewcommand\twocolumn[1][]{#1}%
% \icmltitle{Large-Scale Policy Optimization}
\icmltitle{SAPG: Split and Aggregate Policy Gradients}
% Off
\icmlsetsymbol{equal}{*}

\begin{icmlauthorlist}
\icmlauthor{Jayesh Singla}{equal,yyy}
\icmlauthor{Ananye Agarwal}{equal,yyy}
\icmlauthor{Deepak Pathak}{yyy} \\ 
\vspace{1em}
Carnegie Mellon University
\end{icmlauthorlist}

% \begin{center}
%     \centering
%     \captionsetup{type=figure}
%     \includegraphics[width=\textwidth]{figures/teaser.pdf}
%     \captionof{figure}{\small \textbf{\ours}: We introduce a new class of on-policy RL algorithms that can scale to tens of thousands of parallel environments. In contrast to conventional algorithms such as PPO which learn a single policy across environments leading to wasted environment capacity, our method learns diverse followers in subsets of environments and combines data from them to learn a more optimal leader. Different choices for parametrizing different policies, combining data from them and encouraging diversity lead to different instantiations of our method. We show that \ours scales to upto 24k environments in hard manipulation environments where conventional algorithms fail.}
% \end{center}%

\icmlaffiliation{yyy}{Carnegie Mellon University}
%\icmlaffiliation{comp}{Company Name, Location, Country}
%\icmlaffiliation{sch}{School of ZZZ, Institute of WWW, Location, Country}

\icmlcorrespondingauthor{Ananye Agarwal}{ananyea@andrew.cmu.edu}
\icmlcorrespondingauthor{Jayesh Singla}{jsingla@andrew.cmu.edu}
\icmlkeywords{Machine Learning, ICML}
% \vskip 0.3in
}
{
 \includegraphics[width=\linewidth]{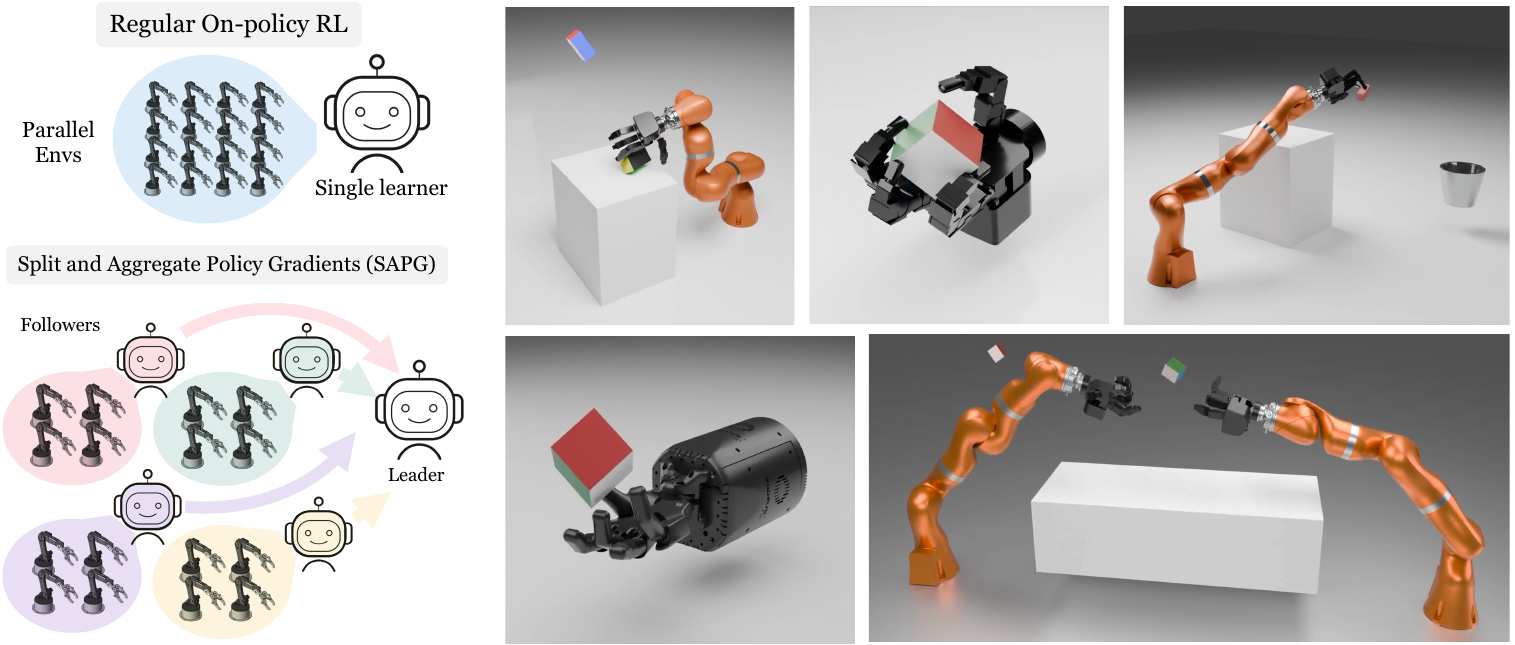}
 \vspace{-0.3in}
 \captionsetup{type=figure}
 \captionof{figure}{\small We introduce a new class of on-policy RL algorithms that can scale to tens of thousands of parallel environments. In contrast to regular on-policy RL, such as PPO, which learns a single policy across environments leading to wasted environment capacity, our method learns diverse followers and combines data from them to learn a more optimal leader in a continuous online manner.}
 \label{fig:teaser}
}\par\bigskip]



\begin{abstract}
Despite extreme sample inefficiency, on-policy reinforcement learning, aka policy gradients, has become a fundamental tool in decision-making problems.
With the recent advances in GPU-driven simulation, the ability to collect large amounts of data for RL training has scaled exponentially. However, we show that current RL methods, e.g. PPO, fail to ingest the benefit of parallelized environments beyond a certain point and their performance saturates. To address this, we propose a new on-policy RL algorithm that can effectively leverage large-scale environments by splitting them into chunks and fusing them back together via importance sampling. Our algorithm, termed SAPG, shows significantly higher performance across a variety of challenging environments where vanilla PPO and other strong baselines fail to achieve high performance. Webpage at \url{https://sapg-rl.github.io}. 

% Although deep reinforcement learning has seen success in many discrete and continuous control tasks, its direct application in hard robotics tasks with a high degree of freedom has been limited, largely due to the high sample complexity of RL techniques. With the advent of highly parallel GPU simulators like IsaacGym, the data throughput available for these algorithms has increased exponentially. Previous works such as \citep{li2023parallel, dexPBT} have explored different methods for scaling off-policy as well as on-policy RL algorithms.  We present a new method to scale 
% PPO \citep{ppo} %on-policy algoriths? 
% aimed at improving usage of collected data and exploration that surpasses these previous methods in terms of amount of experience collected from the environment. Our code will be made available publicly. 
\end{abstract}

\vspace{-2em}
\section{Introduction}
\label{introduction}
% Reinforcement learning (RL) provides a way to generate complex behaviors from simple reward functions. Due to its generality, RL has emerged as a powerful tool to train agents in a variety of environments from video games, long-horizon reasoning, robot manipulation to locomotion. This success has largely been made possible by training either in virtual environments or doing sim2real transfer from physics engines where one could simulate years of real-world experience in minutes to hours.
Broadly, there are two main categories in reinforcement learning (RL): off-policy RL, e.g., Q-learning \citep{Qlearning}, and on-policy RL, e.g., policy gradients \citep{policy_gradient}. On-policy methods are relatively more sample inefficient than off-policy but often tend to converge to higher asymptotic performance. Due to this reason, on-policy RL methods, especially PPO \citep{ppo}, are usually the preferred RL paradigm for almost all sim2real robotic applications \citep{miki2022learning,vision-loco,chen2021general} to games such as StarCraft \citep{starcraft}, where one could simulate years of real-world experience in minutes to hours.

%\begin{figure}[t]
%    \centering
%    \includegraphics[width=\linewidth]{figures/teaser_new.png}
%    \vspace{-0.3in}
%    \caption{\small We introduce a new class of on-policy RL algorithms that can scale to tens of thousands of parallel environments. In contrast to regular on-policy RL, such as PPO, which learns a single policy across environments leading to wasted environment capacity, our method learns diverse followers and combines data from them to learn a more optimal leader in a continuous online manner.}
%    \label{fig:teaser}
%    \vspace{-0.1in}
%\end{figure}

RL is fundamentally a trial-n-error-based framework and hence is sample inefficient in nature. Due to this, one needs to have large batch sizes for each policy update, especially in the case of on-policy methods because they can only use data from current experience. Fortunately, in recent years, the ability to simulate a large number of environments in parallel has become exponentially larger due to GPU-driven physics engines, such as IsaacGym \citep{makoviychuk2021isaac}, PhysX, Mujoco-3.0, etc. This means that each RL update can easily scale to batches of size hundreds of thousands to millions, which are over two orders of magnitude higher than what most RL benchmarks typically have.

\begin{figure}[t]
    \centering
    \includegraphics[width=\linewidth]{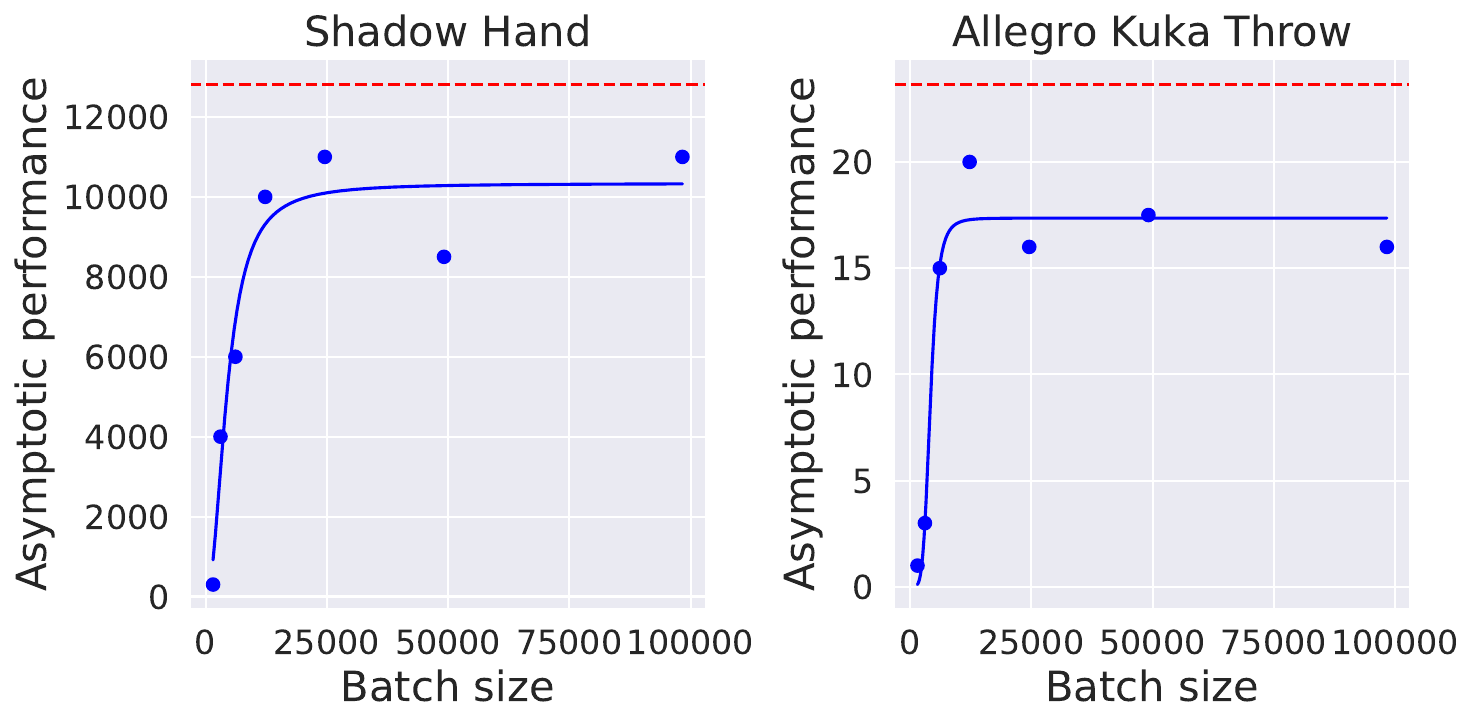}
    \vspace{-2.2em}
    \caption{Performance vs batch size plot for PPO runs (blue curve) across two environments. The curve shows how PPO training runs can not take benefit of large batch size resulting from massively parallelized environments and their asymptotic performance saturates after a certain point. The dashed red line is the performance of our method, SAPG, with more details in the results section. It serves as evidence that higher performance is achievable with larger batch sizes.}
    \label{fig:motivate}
    \vspace{-1em}
\end{figure}

In this paper, we highlight an issue with typical on-policy RL methods, e.g. PPO, that they are not able to ingest the benefits with increasingly larger sample sizes for each update. In Figure~\ref{fig:motivate}, we show that PPO performance saturates after a certain batch size despite the ceiling being higher. This is due to the issue in data sampling mechanisms. In particular, at each timestep actions are sampled from a Gaussian with some mean and variance. This implies that most sampled actions are near the mean and with large number of environments, many environments are executing the same actions leading to duplicated data. This implies that the performance of PPO saturates at some point as we increase the number of environments. 

We propose a simple fix to this problem. Instead of running a single PPO policy for all environments, we divide environments into blocks. Each block optimizes a separate policy, allowing for more data diversity than just i.i.d. sampling from the same Gaussian. Next, we do an off-policy update to combine data from all these policies to keep the update consistent with the objective of on-policy RL. This allows us to use the PPO's clipped surrogate objective, maintaining the stability benefits of PPO while latching onto high reward trajectories even though they are off-policy. A schematic of our approach, termed SAPG, is shown in Figure~\ref{fig:teaser}. We evaluate SAPG across a variety of environments and show significantly high asymptotic performance in environments where vanilla PPO even fails to get any positive success.

\section{Related Work}

\paragraph{Policy gradients}
% Policy gradient algorithms have been one of the highly used methods in reinforcement learning. The core idea is to update the policy using some form of gradient ascent on an estimator of the gradient of the reward objective with respect to the policy. 
REINFORCE \citep{reinforce}, one of the earliest policy gradient algorithms uses an estimator of the objective using simple Monte Carlo return values. Works such as \citep{actorcritic} and \citep{gae} improve the stability of policy gradient algorithms by employing a baseline to decrease the variance of the estimator while not compromising on the bias. \citep{trpo,ppo} incorporate conservative policy updates into policy gradients to increase the robustness.
% by preventing large updates in the policy space.

\paragraph{Distributed reinforcement learning} %Before the introduction of GPU based simulation, previous work sped up RL training by distributing data collection/training across multiple processes. GORILA \citep{gorila} builds upon the DQN architecture and separates acting and learning into different processes. Ape-X \citep{apex} provide a framework for distributed prioritized replay \citep{per} to be implemented over off-policy learning methods. For on-policy algorithms, \citep{a3c}, \citep{impala} introduce asynchronous SGD based policy gradient methods. \citep{dppo} introduce Distributed PPO, showing increased training speed on locomotion tasks. 
Reinforcement learning algorithms are highly sample inefficient, which calls for some form of parallelization to increase the training speed. When training in simulation, this speed-up can be achieved by distributing experience collection or different parts of training across multiple processes.\citep{gorila, a3c, impala, apex}
However, through the introduction of GPU-based simulators such as IsaacGym \citep{makoviychuk2021isaac}, the capacity of simulation has increased by two to three orders of magnitude. 
% This massive shift has completely changed the landscape of distributed reinforcement learning.
Due to this, instead of focusing on how to parallelize parts of the algorithm, the focus has shifted to finding ways to efficiently utilize the large amount of simulation data.
Previous works such as \citep{vision-loco, chen2021general, Fu2022DeepWC, rudin2022learning, Handa2022DeXtremeTO} use data from GPU-based simulation to learn policies in complex manipulation and locomotion settings. However, most of these works still use reinforcement learning algorithms to learn a single policy, while augmenting training with techniques like teacher-student-based training and game-based curriculum. We find that using the increased simulation capacity to naively increase the batch size is not the best way to utilize massively parallel simulation. 
%This is because the reduction in sample variance due to the increase in batch size will be insignificant to training after a point.

\citep{dexPBT} develop a population-based training framework that divides the large number of environments between multiple policies and using hyperparameter mutation to find a set of hyperparameters that performs well. However, even this does not utilize all the data completely as each policy learns independently. We propose a way to ensure most of the data from the environments contributes to learning by using all collected transitions for each update.
\paragraph{Off-policy Policy Gradients}
Unlike on-policy algorithms, off-policy algorithms can reuse all collected data or data collected by any policy for their update. Most off-policy algorithms \citep{dqn, ddpg, sac} try to learn a value function which is then implicitly/explicitly used to learn a policy.
\citep{li2023parallel} developed a variant of Deep Deterministic Policy Gradient (DDPG) called PQL which splits data collection and learning into multiple processes and shows impressive performance on many benchmark tasks. We use PQL as one of our baselines to compare our method to off-policy RL in complex tasks.
Although off-policy algorithms are much more data-efficient, they usually get lower asymptotic performance than on-policy policy gradients. This has inspired works to develop techniques to use off-policy data in on-policy methods. \citep{imp_sampling} has been one of the major techniques used to realize this. Previous works \citep{offp_actor_critic, acer, impala, p3o} develop techniques to use off-policy data in on-policy algorithms using importance sampling-based updates along with features such as bias correction.
% Our method presents a new way to use off-policy updates to 

% ==========================================
% ==========================================

\section{Preliminaries}
In this paper, we propose a modification to on-policy RL to achieve higher performance in the presence of large batch sizes. We build upon PPO, although our proposed ideas are generally applicable to any on-policy RL method.

\paragraph{On-policy RL}
Let ($\mathcal{S}, \mathcal{A}, \mathcal{P}, r, \rho, \gamma)$ be an MDP where $\mathcal{S}$ is the set of states, $\mathcal{A}$ the set of actions, $\mathcal{P}$ are transition probabilities, $r$ the reward function, $\rho$ the initial distribution of states and $\gamma$ the discount factor. The objective in reinforcement learning is to find a policy $\pi(a|s)$ which maximises the long term discounted reward $\mathcal{J}(\pi) = \mathop{\mathbb{E}}_{s_0 \sim \rho, a_t \sim \pi(\cdot|s_t) }\left[ \sum_{t=0}^{T-1} \gamma^tr(s_t, a_t)\right]$. 

Policy-gradient algorithms \citep{reinforce, actorcritic, trpo, a3c} optimize the policy using gradient descent with Monte Carlo estimates of the gradient
\begin{equation}
\nabla_{\theta} J(\pi_\theta) = \mathop{\mathbb{E}}_{s \sim \rho_d, a \sim \pi(\cdot| s)}\left[\nabla_{\theta}\log(\pi_\theta(a))\hat{A}^{\pi_\theta}(s,a)\right] 
\label{eq:actor_critic}
\end{equation}

where $\hat{A}^{\pi_\theta}(s,a)$ is an advantage function that estimates the contribution of the transition to the gradient. A common choice is $\hat{A}^{\pi_\theta}(s,a) = \hat{Q}^{\pi_\theta}(s,a) - \hat{V}^{\pi_{\theta}}(s)$, where $\hat{Q}^{\pi_\theta}(s,a)$, $\hat{V}^{\pi_{\theta}}(s)$ are estimated $Q$ and value functions. This form of update is termed as an actor-critic update \citep{actorcritic}. Since we want the gradient of the error with respect to the current policy, only data from the current policy (on-policy) data can be utilized. 

\paragraph{PPO} Actor critic updates can be quite unstable because gradient estimates are high variance and the loss landscape is complex. An update step that is too large can destroy policy performance. Proximal Policy Optimization (PPO) modifies Eq.~\ref{eq:actor_critic} to restrict updates to remain within an approximate ``trust region" where there is guaranteed improvement \citep{trpo, kakade2002approximately}. 
% \begin{equation}
% \begin{split}
%     L_{on}(\pi_\theta) = \mathop{\mathbb{E}}_{\pi_{old}}\left[ \min(r_t(\pi_\theta), 
%     & \mathop{\mathrm{clip}}(r_t(\pi_\theta), 1 - \epsilon, \right. \\ 
%     & \left. 1+ \epsilon))A_t^{\pi_{old}}\right]  
% \end{split}
% \label{eq:ppo_loss}
% \end{equation}

\begin{equation}
\begin{split}
    L_\mathrm{on}(\pi_\theta) &= \mathop{\mathbb{E}}_{\pi_\mathrm{old}}\left[ \min(r_t(\pi_\theta), \right. \\
    & \left. \mathop{\mathrm{clip}}\left(r_t(\pi_\theta), 1 - \epsilon, 1+ \epsilon\right))A_t^{\pi_{old}}\right]  
\end{split}
\label{eq:ppo_loss}
\end{equation}

Here, $r_t(\pi_\theta) = \frac{\pi_{\theta}(a_t|s_t)}{\pi_{old}(a_t|s_t)}$, $\epsilon$ is a clipping hyperparameter and $\pi_{old}$ is the policy collecting the on-policy data. The clipping operation ensures that the updated $\pi$ stays close to $\pi_{old}$. Empirically, given large numbers of samples, PPO achieves high performance, is stable and robust to hyper-parameters. However, it was developed for relatively small batch sizes ($\approx 100$ parallel envs). We find that in the large-scale setting ($>$10k envs), it is suboptimal because many parallel envs are sampling nearly identical on-policy data. 

\begin{figure*}
    \centering
    \includegraphics[width=\linewidth]{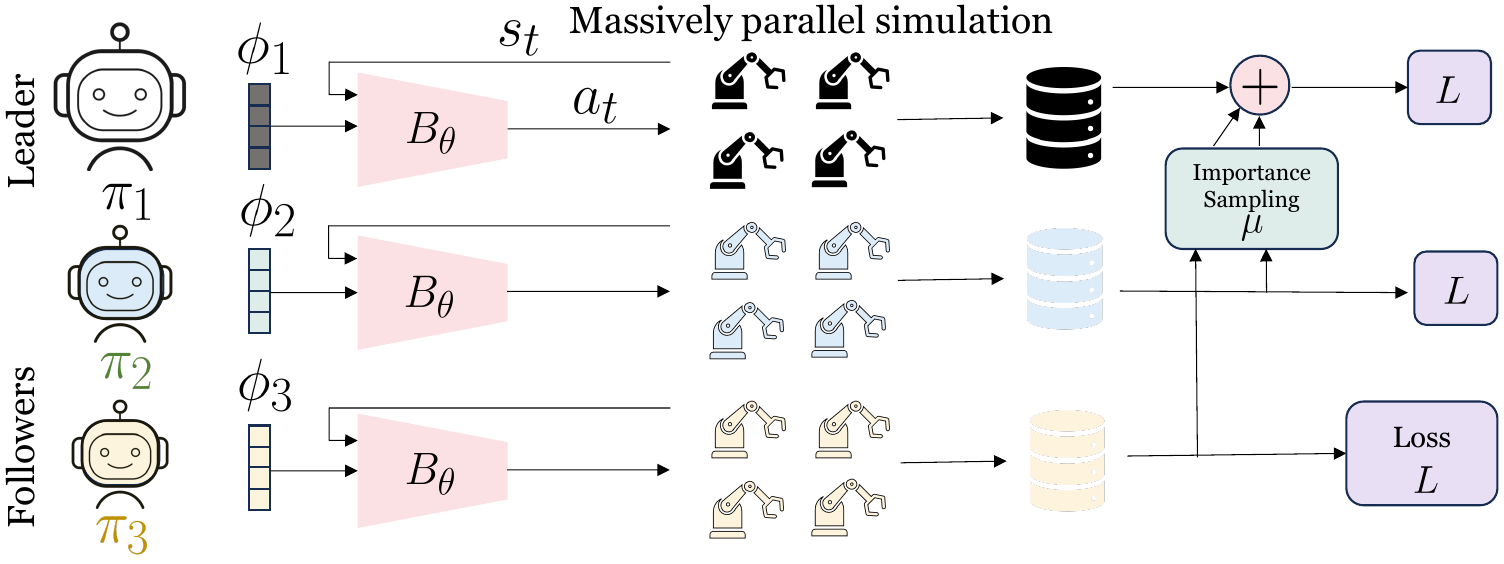}
    \vspace{-2em}
    \caption{We illustrate one particular variant of \ours which performs well. There is one leader and $M-1$ followers ($M=3$ in figure). Each policy has the same backbone with shared parameters $B_\theta$ but is conditioned on local learned parameters $\phi_i$. Each policy gets a block     of $\frac{N}{M}$ environments to run. The leader is updated with its on-policy data as well as importance-sampled off-policy data from the followers. Each of the followers only uses their own data for on-policy updates.}
    \label{fig:method}
\end{figure*}

\section{Split and Aggregate Policy Gradients}
Policy gradient methods are highly sensitive to the variance in the estimate of gradient. Since CPU-based simulators typically run only 100s of environments in parallel, conventional wisdom is to simply sample on-policy data from a Gaussian policy in all the environments since as the number of datapoints increases, the Monte Carlo estimate becomes more accurate. However, this intuition no longer holds in the extremely large-scale data setting where we have hundreds of thousands of environments on GPU-accelerated simulators like IsaacGym. IID sampling from a Gaussian policy will lead to most actions lying near the mean, and most environments will execute similar actions, leading to wasted data (fig.~\ref{fig:motivate}). 

We propose to efficiently use large numbers of $N$ environments using a divide-and-conquer setup. Our algorithm trains a variety of $M$ policies $\pi_1, \ldots, \pi_M$ instead of having just one policy. However, simply training multiple policies by dividing environments between them is also inefficient. This is equivalent to training an algorithm with different seeds and choosing the best seed. One approach is to add hyperparameter mutation \citep{dexPBT} to the policies and choosing the hyperparameters that perform the best among all of them. However, even in this case, all of the data from the ``worse" policies goes to waste, and the only information gained is that some combinations of hyperparameters are bad, even though the policies themselves may have discovered high reward trajectories. We need to somehow aggregate data from multiple policies into a single update. We propose to do this via off-policy updates. 

\subsection{Aggregating data using off-policy updates}
One of the major drawbacks of on-policy RL is its inability to use data from past versions of the policy. One solution is to use importance sampling \citep{offp_actor_critic, Meng2023} to weight updates using data from different policies. In practice, this is not used since given limited compute it is beneficial to sample on-policy experience that is more directly relevant. However, this is no longer true in the large batch setting where enough on-policy data is available. In this case, it becomes advantageous to have multiple policies $\pi_1, \ldots, \pi_M$ and use them to sample diverse data, even if it is off-policy. In particular, to update policy $\pi_i$ using data from policy $\pi_j$, $j\in \mathcal{X}$ we use \citep{Meng2023}
\begin{equation}
    \begin{split}
    & L_\textrm{off}(\pi_i; \mathcal{X}) = \frac{1}{|\mathcal{X}|}\sum_{j\in\mathcal{X}}\mathop{\mathbb{E}}_{(s, a) \sim \pi_j}\left[\min\left(r_{\pi_i}(s, a), \right. \right. \\
    & \left. \left. \mathop{\mathrm{clip}}\left(r_{\pi_i}(s, a), \mu(1 - \epsilon), \mu(1 + \epsilon)\right)\right)A^{\pi_{i, \textrm{old}}}(s, a)\right]
    \label{eq:offp_ppo}
\end{split}
\end{equation}
where $r_{\pi_i}(s, a) = \frac{\pi_i(s, a)}{\pi_j(s, a)}$ and $\mu$ is an off-policy correction term $\mu = \frac{\pi_{i, old}(s, a)}{\pi_j(s,a)}$. Note that when $i=j$, then $\pi_j = \pi_{i, old}$ and this reduces to the on-policy update as expected. This is then scaled and combined with the on-policy term (eq.~\ref{eq:ppo_loss})
\begin{equation}
    L(\pi_i) = L_{on}(\pi_i) + \lambda \cdot L_{off} (\pi_i; \mathcal{X})
\end{equation}
The update target for the critic is calculated using $n$-step returns (here $n=3$). 
\begin{equation}
    V_{on, \pi_j}^{target}(s_t) = \sum_{k=t}^{t+2} \gamma^{k-t}r_k + \gamma^3 V_{\pi_j, old}(s_{t+3})
\end{equation}
However, this is not possible for off-policy data. Instead, we assume that an off-policy transition can be used to approximate a 1-step return. The target equations are as follows - 
\begin{equation}
    V_{off, \pi_j}^{target}(s'_t) = r_t + \gamma V_{\pi_j, old}(s'_{t+1})
\end{equation}
The critic loss is then
\begin{equation}
\label{eq:onp_critic_loss}
    L_{on}^{critic}(\pi_i) = \mathop{\mathbb{E}}_{(s, a) \sim \pi_i}\left[(V_{\pi_i}(s) - V_{on, \pi_i}^{target}(s))^2\right] \\
\end{equation}
\begin{equation}
    L_{off}^{critic}(\pi_i;\mathcal{X})  = \frac{1}{|\mathcal{X}|} \sum_{j \in \mathcal{X}} \mathop{\mathbb{E}}_{(s, a) \sim \pi_j}\left[(V_{\pi_i}(s) - V_{off, \pi_i}^{target}(s))^2\right]
\label{eq:offp_critic_loss}
\end{equation}
\begin{equation}
\label{eq:net_critic_loss}
    L^{critic}(\pi_i) = L_{on}^{critic}(\pi_i) + \lambda \cdot L_{off}^{critic}(\pi_i) 
\end{equation}
Given this update scheme, we must now choose a suitable $\mathcal{X} \subseteq \{1, \ldots, M\}$ and the set of $i$s to update, along with the correct ratio $\lambda$. We explore several variants below. 

\subsection{Symmetric aggregation}
\label{sec:symmetric_agg}
A simple choice is to update all $i's$ with the data from all policies. In this case, we choose to update each policy $i\in\{1, 2, \ldots, M\}$ and for each $i$ use off-policy data from all other policies $\mathcal{X} = \{1, 2, i - 1, i + 1, \ldots, M\}$. Since gradients from off-policy data are typically noisier than gradients from on-policy data, we choose $\lambda = 1$, but sub-sample the off-policy data such that we use equal amounts of on-policy and off-policy data.  

\subsection{Leader-follower aggregation}
While the above choice prevents data wastage, since all the policies are updated with the same data, it can lead to policies converging in behavior, reducing data diversity and defeating the purpose of having separate policies. To resolve this, we break symmetry by designating a ``leader" policy $i=1$ which gets data from all other policies $\mathcal{X} = \{2, 3, \ldots, M\}$ while the rest are ``followers" and only use their own on-policy data for updates $\mathcal{X} = \phi$. As before, we choose $\lambda = 1$, but subsample the off-policy data for the leader such that we use equal amounts of on-policy and off-policy data in a mini-batch update.

\subsection{Encouraging diversity via latent conditioning} 
What is the right parameterization for this set of policies? One simple choice is to have a disjoint set of parameters for each with no sharing at all. However, this implies that      each follower policy has no knowledge of any other policy whatsoever and may get stuck in a bad local optimum. We mitigate this by having a shared backbone $B_\theta$ for each policy conditioned on hanging parameters $\phi_j$ local to each policy. Similarly, the critic consists of a shared backbone $C_\psi$ conditioned on parameters $\phi_j$. The parameters $\psi, \theta$ are shared across the leader and all followers and updated with gradients from each objective, while the parameters $\phi_j$ are only updated with the objective for that particular policy. We choose $\phi_j \in \mathbb{R}^{32}$ for complex environments while $\phi_j\in\mathbb{R}^{16}$ for the relatively simpler ones.  

\subsection{Enforcing diversity through entropy regularization}
To further encourage diversity between different policies, in addition to the PPO update loss $L_{on}$ we add an entropy loss to each of the followers with different coefficients. In particular, the entropy loss is $\mathcal{H}(\pi(a\mid s))$. The overall loss for the policy $i$ (or the $(i-1)$th) follower is $L(\pi_i) = L_{on}(\pi_i) + \lambda_\textrm{ent}(i-1)\cdot\mathcal{H}(\pi(a\mid s))$. The leader doesn't have any entropy loss. Different scales of coefficients produce policies with different explore-exploit tradeoffs. Followers with large entropy losses tend to explore more actions even if they are suboptimal, while those with small coefficients stay close to optimal trajectories and refine them. This leads to a large data coverage with a good mix of optimal as well as diverse trajectories. We treat $\lambda_\textrm{ent}$ as a hyperparameter. 
\begin{figure}[ht!]
    \centering
    \includegraphics[width=\linewidth]{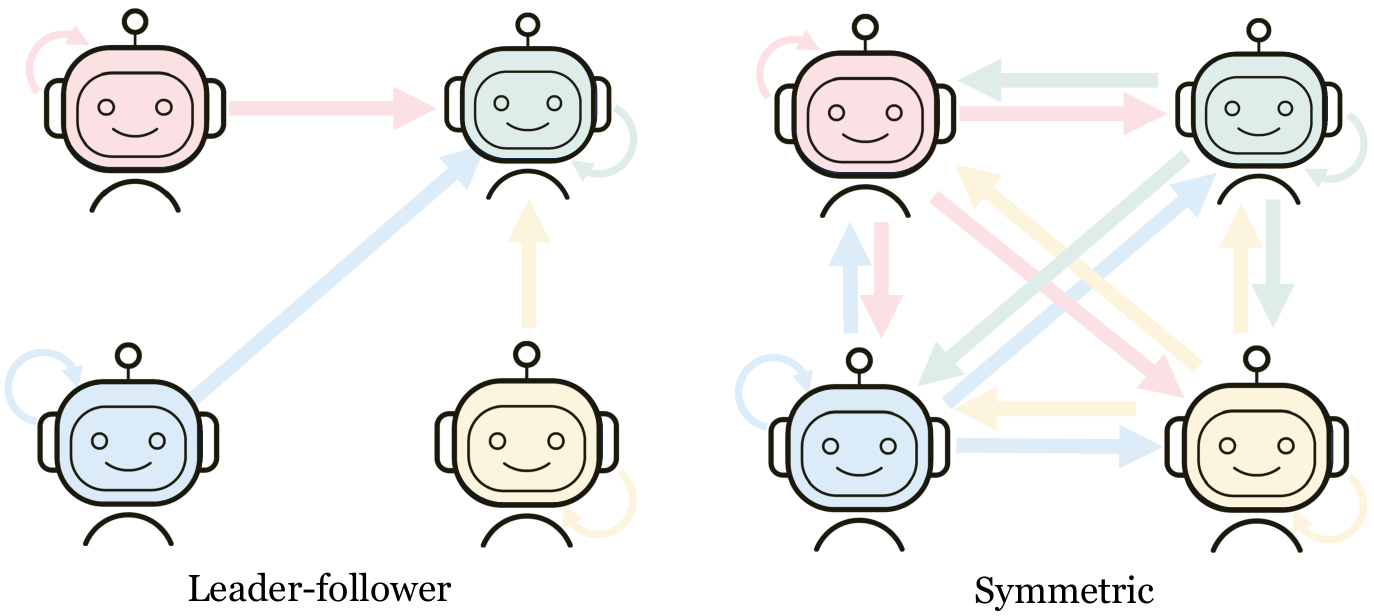}
    
    \caption{Two data aggregation schemes we consider in this paper. (Left) one policy is a leader and uses data from each of the followers (Right) a symmetric scheme where each policy uses data from all others. In each case, the policy also uses its own on-policy data.}
    \label{fig:aggregation}
\end{figure}
\vspace{-1.5em}
\subsection{Algorithm: SAPG}

We roll out $M$ different policies and collect data $\mathcal{D}_1, \ldots, \mathcal{D}_M$ for each. Follower policies $2, \ldots, M$ are updated using the usual PPO objective with minibatch gradient descent on their respective datasets. However, we augment the dataset of the leader $\mathcal{D}_1$ with data from $\mathcal{D}_2, \ldots, \mathcal{D}_M$, weighed by the importance weight $\mu$. The leader is then updated by minibatch gradient descent as well. 

\begin{algorithm}[t]
\caption{\ours}
\begin{algorithmic}
\STATE Initialize shared parameters $\theta$, $\psi$ 
\STATE For $i\in\{1, \ldots, M\}$ initialize parameters $\phi_i$
\STATE Initialize $N$ environments $E_1, \ldots, E_N$. 
\STATE Initialize data buffers for each policy $\mathcal{D}_1, \ldots, \mathcal{D}_M$. 
\FOR{$i=1, 2, \ldots, $}
\FOR{$j=1, 2, \ldots, M$}
\STATE $\mathcal{D}_j \leftarrow$ CollectData$\left(E_{j\frac{N}{M}:(j+1)\frac{N}{M}}, \theta, \psi_j\right)$
\ENDFOR
\STATE $L \leftarrow 0$ 
\STATE Sample $|\mathcal{D}_1|$ transitions from $\mathop{\cup}_{j=2}^M \mathcal{D}_i$ to get $\mathcal{D}_1'$.
\STATE $L \leftarrow L$ + OffPolicyLoss$\left(\mathcal{D}_1'\right)$ 
\STATE $L \leftarrow L$ + OnPolicyLoss$\left(\mathcal{D}_1\right)$
\FOR{$j=2, \ldots, M$}
\STATE $L \leftarrow L$ + OnPolicyLoss{$\left(\mathcal{D}_j\right)$} 
\ENDFOR
\STATE Update $\theta \leftarrow \theta - \eta \nabla_\theta L$
\STATE Update $\psi \leftarrow \psi - \eta \nabla_\psi L$
\ENDFOR
\end{algorithmic}
\label{alg:sapg}
\end{algorithm}

\section{Experimental Setup}
We conduct experiments on 5 manipulation tasks (3 hard and 2 easy) and compare them against SOTA methods for the large-scale parallelized setting. We use a GPU-accelerated simulator, IsaacGym \citep{makoviychuk2021isaac} which allows simulating tens of thousands of environments in parallel on a single GPU. In our experiments, we focus on the large-scale setting and simulate $24576$ parallel environments unless otherwise specified. Note that this is \textbf{two orders of magnitude} larger than the number of environments PPO \citep{ppo} was developed on, and we indeed find that vanilla PPO does not scale to this setting. 

For testing, we choose a suite of manipulation environments that are challenging and require large-scale data to learn effective policies \citep{dexPBT}. In particular, these consist of dexterous hands mounted on arms leading to high numbers of degrees of freedom (up to 23). This is challenging because sample complexity scales exponentially with degrees of freedom. They are under-actuated and involve manipulating free objects in certain ways while under the influence of gravity. This leads to complex, non-linear interactions between the agent and the environment such as contacts between robot and object, object and table, and robot and table. Overall, this implies that to learn effective policies an agent must collect a large amount of relevant experience and also use it efficiently for learning.

\subsection{Tasks}
We consider a total of 6 tasks grouped into two parts: Four hard tasks and two easy tasks. Hard and easy is defined by the success reward achieved by off-policy (in particular, PQL) methods in these environments. In easy environments, even Q-learning-based off-policy methods can obtain non-zero performance but not in hard tasks. See appendix sec.~\ref{appendix:env}.

\paragraph{Hard Difficulty Tasks}
All four hard tasks are based on the \texttt{Allegro-Kuka} environments \citep{dexPBT}. These consist of an Allegro Hand (16 DoF) mounted on a Kuka arm (7 dof). The performance of the agent in the above three tasks is measured by the successes metric which is defined as the number of successes in a single episode. Three tasks include:
\begin{itemize}[noitemsep,left=0pt,topsep=-1pt]
    \item \textbf{Regrasping}: The object must be lifted from the table and held near a goal position $\mathbf{g}_t \in \mathbb{R}^3$ for $K=30$ steps. This is called a ``success". The target position and object position are reset to a random location after every success.
    \item \textbf{Throw}: The object must be lifted from the table and thrown into a bucket at $\mathbf{g}_t\in\mathbb{R}^3$ placed out of reach of the arm. The bucket and the object position are reset randomly after every successful attempt.
    \item \textbf{Reorientation}: Pick up the object and reorient it to a particular target pose $\mathbf{g_t}\in\mathbb{R}^7$ (position + orientation). The target pose is reset once the agent succeeds. This means that the agents needs to reorient the object in different poses in succession, which may sometimes entail placing the objects on the table and lifting it up in a different way.

    \item \textbf{Two Arms Reorientation}: Similar to the reorientation task above, pick up the object and reorient it to a particular target pose. However, there are two arms in the system, adding the additional complexity of having to transfer objects between arms to reach poses in different regions of space.
    
\end{itemize} 

\begin{figure*}[t]
    \centering
    \includegraphics[width=\textwidth,height=8cm]{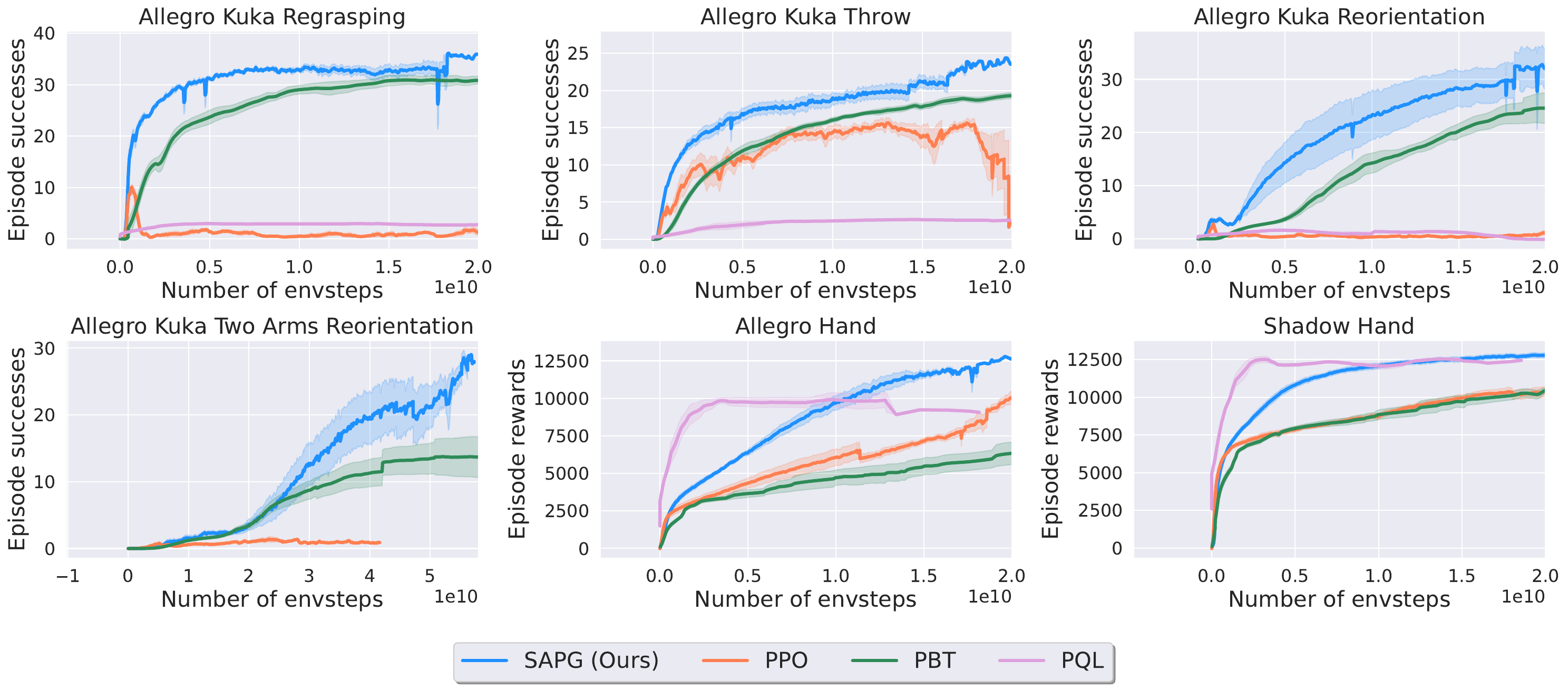}
    \caption{Performance curves of \ours with respect to PPO, PBT and PQL baselines. On AllegroKuka tasks, PPO and PQL barely make progress and \ours beats PBT. On Shadow Hand and Allegro Kuka Reorientatio and Two Arms Reorientation, \ours performs best with an entropy coefficient of 0.005 while the coefficient is 0 for other environments. On ShadowHand and AllegroHand, while PQL is initially more sample efficient, \ours is more performant in the longer run. AllegroKuka environments use successes as a performance metric while AllegroHand and ShadowHand use episode rewards.}
    \label{fig:result}
    \vspace{-1em}
\end{figure*}

\paragraph{Easy Difficulty Tasks:}
In addition, we test on the following dexterous hand tasks. As before, the observation space consists of the joint angles and velocities $\mathbf{q}_t$, $\dot{\mathbf{q}}_t$, object pose $\mathbf{x}_t$ and velocities $\mathbf{v}_t, \mathbf{\omega}_t$.  
\begin{itemize}[noitemsep,left=0pt,topsep=-1pt]
    \item \textbf{Shadow Hand}: We test in-hand reorientation task of a cube using the 24-DoF Shadow Hand. 
    \item \textbf{Allegro Hand}: This is the same as the previous in-hand reorientation task but with the 16-DoF Allegro Hand. 
\end{itemize}

\subsection{Baselines}
\label{sec:baselines}
We test against state-of-the-art RL methods designed for the GPU-accelerated large-scale setting we consider in this paper. We compare against both on-policy \citep{dexPBT} and off-policy \citep{li2023parallel} variants as well as vanilla PPO \citep{ppo}. 
\begin{itemize}[noitemsep,left=0pt,topsep=-1pt]
    \item \textbf{PPO} (Proximal Policy Optimization) \citep{ppo}: In our setting, we just increase the data throughput for PPO by increasing the batch size proportionately to the number of environments. In particular, we see over two orders of magnitude increase in the number of environments (from 128 to 24576).   
     \item \textbf{Parallel Q-Learning} \citep{li2023parallel} A parallelized version of DDPG with different mixed exploration i.e. varying exploration noise across environments to further aid exploration. We use this baseline to compare if off-policy methods can outperform on-policy methods when the data collection capacity is high. 
    \item \textbf{DexPBT} \citep{dexPBT} A framework that combines population-based training with PPO. $N$ Environments are divided into $M$ groups, each containing $\frac{N}{M}$ environments. $M$ separate policies are trained using PPO in each group of environments with different hyperparameters. At regular intervals, the worst-performing policies are replaced with the weights of best-performing policies and their hyperparameters are mutated randomly.
    % The idea is to use the increased simulation capacity of IsaacGym search over a large set of training as well as reward hyperparameters. Note that PBT can mutate the reward scales to directly optimize the success rate, whereas our method is not allowed to do so and must optimize the auxiliary reward function objective.
\end{itemize}

Due to the complexity of these tasks, experiments take about 48-60 hours on a single GPU, collecting $\approx 2\mathrm{e}{10}$ transitions. Since we run experiments on different machines, the wall clock time is not directly comparable and we compare runs against the number of samples collected. We run 5 seeds for each experiment and report the mean and standard error in the plots. In each plot, the solid line is $y(t) = \frac{1}{n}\sum_i y_i(t)$ while the width of the shaded region is determined by standard error $\frac{2}{\sqrt{n}}\sum_i (y(t) - y_i(t))^2$.

For each task, we use $M=6$ policies for our method and DexPBT in a total of $N=24576$ environments for each method. We use the dimension of learned parameter $\phi_j\in\mathbb{R}^{32}$ for the AllegroKuka tasks while we use $\phi_j\in\mathbb{R}^{16}$ for the ShadowHand and AllegroHand tasks since they are relatively simpler. We use a recurrent policy for the AllegroKuka tasks and an MLP policy for the Shadow Hand and Allegro Hand tasks and use PPO to train them. We collect 16 steps of experience per instance of the environment before every PPO update step. For \ours, we tune the entropy coefficient $\sigma$ by choosing the best from a small set $\{0, 0.003, 0.005\}$ for each environment. We find that $\sigma=0$ works best for all AllegroHand, Regrasping, and Throw while $\sigma=0.005$ works better for ShadowHand and Reorientation.   

\begin{table*}[t]
\vskip 0.15in
\begin{center}
\begin{small}
\begin{sc}
\begin{tabularx}{\textwidth}{l*{5}{>{\centering\arraybackslash}X}}
\toprule
Task & PPO~\citep{ppo} & PBT~\citep{dexPBT} & PQL~\citep{li2023parallel} & SAPG ($\lambda_\textrm{ent}=0$) & SAPG ($\lambda_\textrm{ent}=0.005$) \\
\midrule
AllegroHand & $1.01 \mathrm{e}4 \pm 6.31\mathrm{e}2$ & $7.28 \mathrm{e}3 \pm 1.24\mathrm{e}3$ & $1.01 \mathrm{e}4 \pm 5.28\mathrm{e}2$ & $\mathbf{1.23 \mathrm{e}4 \pm 3.29\mathrm{e}2}$ & $9.14 \mathrm{e}3 \pm 8.38\mathrm{e}2$ \\
ShadowHand & $1.07 \mathrm{e}4 \pm 4.90\mathrm{e}2$ & $1.01 \mathrm{e}4 \pm 1.80\mathrm{e}2$ & $1.28 \mathrm{e}4 \pm 1.25\mathrm{e}2$ & $1.17 \mathrm{e}4 \pm 2.64\mathrm{e}2$ & $\mathbf{1.28 \mathrm{e}4 \pm 2.80\mathrm{e}2}$ \\
Regrasping & $1.25 \pm 1.15$ & $31.9 \pm 2.26$ & $2.73 \pm 0.02$ & $\mathbf{35.7 \pm 1.46}$ & $33.4 \pm 2.25$ \\
Throw & $16.8 \pm 0.48$ & $19.2 \pm 1.07$ & $2.62 \pm 0.08$ & $\mathbf{23.7 \pm 0.74}$ & $18.7 \pm 0.43$ \\
Reorientation & $2.85 \pm 0.05$ & $23.2 \pm 4.86$ & $1.66 \pm 0.11$ & $33.2 \pm 4.20$ & $\mathbf{38.6 \pm 0.63}$ \\
Two Arms Reorientation & $1.73 \pm 0.51$ & $14.46 \pm 2.91$ & - & - & $\mathbf{28.58 \pm 1.55}$ \\
\bottomrule
\end{tabularx}
\end{sc}
\end{small}
\end{center}
\vskip -1em
\caption{Performance after $2\mathrm{e}10$ samples for different methods with standard error. This is measured by successes for the AllegroKuka tasks and by episode rewards for in-hand reorientation tasks. Across environments, we find that our method performs better than baselines.}
\end{table*}

% \begin{table*}[t]
% \vskip 0.15in
% \begin{center}
% \begin{small}
% \begin{sc}
% \begin{tabular}{lccccr}
% \toprule
% Algorithm & Allegro Hand & Shadow Hand & Regrasping & Throw & Reorientation \\
% \midrule
% PPO                      & $10063.78 \pm 631.39$           & $10712.49 \pm 489.59$          & $1.25 \pm 1.15$        & $16.81 \pm 0.48 $        & $2.85 \pm 0.05$\\
% DexPBT                   & $7282.88 \pm 1235.28$            & $10144.78 \pm 180.31$         & $31.90 \pm 2.26$       & $19.23 \pm 1.07$          & $23.22 \pm 4.86$\\
% PQL                      & $10138.45 \pm 528.04$           & $12783.32 \pm 124.59$          & $2.73 \pm 0.02$          & $2.62 \pm 0.08$           & $1.66 \pm 0.11$ \\
% \midrule
% Ours (w/o entropy loss)  &  $\mathbf{12338.79 \pm 328.88}$ & $11714.78 \pm 264.22$         & $\mathbf{35.72 \pm 1.46}$ & $\mathbf{23.66 \pm 0.74}$ & $33.23 \pm 4.20$         \\
% Ours (with entropy loss) & $9140.51 \pm 838.14$          & $\mathbf{12819.52 \pm 279.69}$ & $33.41 \pm 2.25$          & $18.65 \pm 0.43$          & $\mathbf{38.63 \pm 0.63}$ \\
% \bottomrule
% \end{tabular}
% \end{sc}
% \end{small}
% \end{center}
% \vspace{-1em}
% \caption{Performance after $2\mathrm{e}10$ samples for different methods with standard error. This is measured by successes for the AllegroKuka tasks and by episode rewards for in-Hand reorientation tasks. Across environments, we find that \ours performs better than baselines}
% \end{table*}

\begin{figure*}[t]
    \centering
    \includegraphics[width=\textwidth,height=8cm]{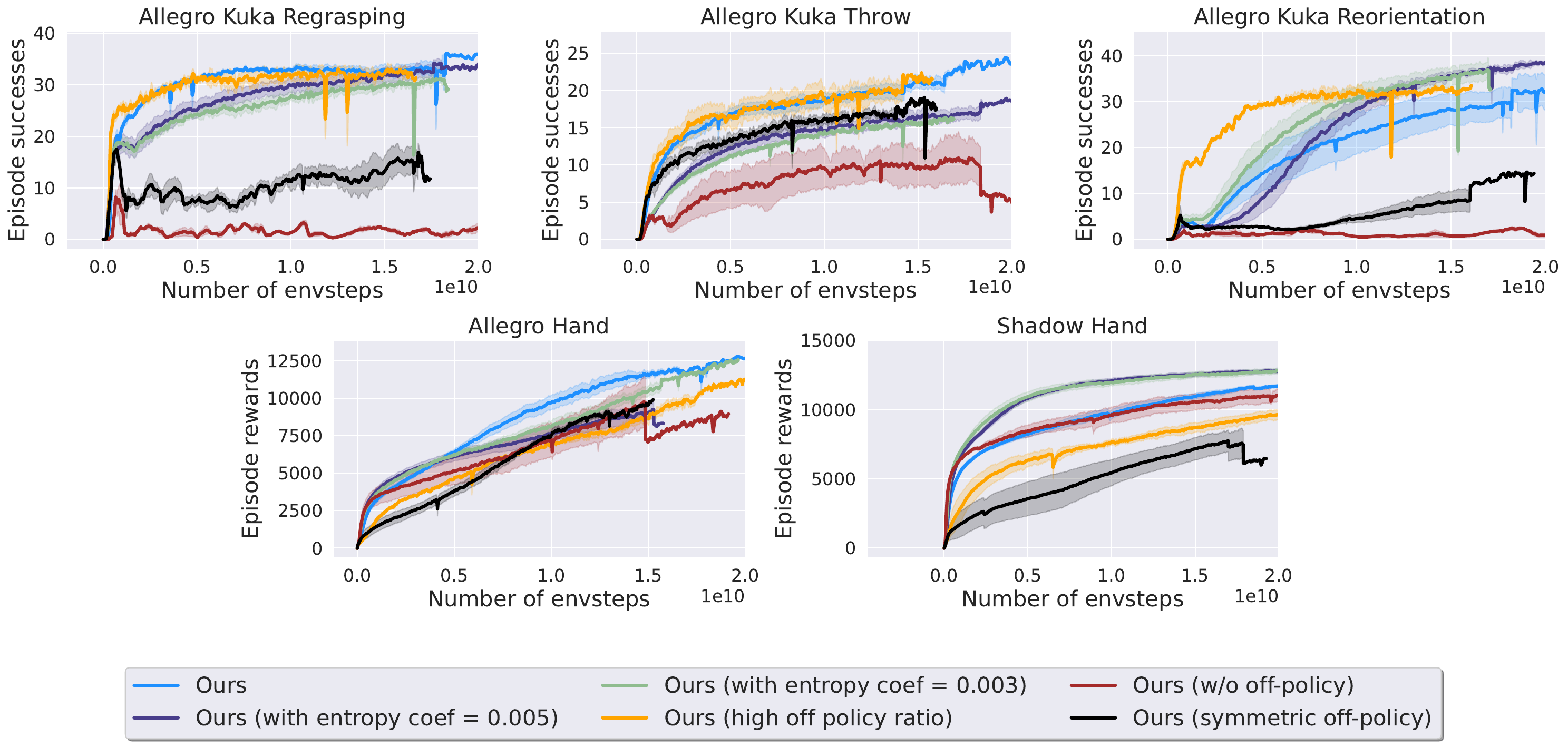}

    \caption{Performance curves for ablations of our method. The variants of our method with a symmetric aggregation scheme or without an off-policy combination perform significantly worse. Entropy regularization affects performance across environments, giving a benefit in reorientation. Using a high off-policy ratio without subsampling data leads to worse performance on ShadowHand and AllegroHand.}
    \label{fig:ablation}
    % \vspace{-1em}
\end{figure*}

\section{Results and Analysis}
In the large-scale data setting, we are primarily concerned with optimality while sample-efficiency and wall-clock time are secondary concerns. This is because data is readily available---one only needs to spin up more GPUs, what is really important is how well our agent performs in the downstream tasks. Indeed, this aligns with how practitioners use RL algorithms in practice \citep{vision-loco, extreme-parkour, openai_shadow_hand}, where agents are trained with lots of domain randomization in large simulations and the primary concern is how well the agent can adapt and learn in these environments since this directly translates to real-world performance.    
\subsection{AllegroKuka tasks}
The AllegroKuka tasks (Throw, Regrasping, Reorientation, Two Arms Reorientation) are hard due to large degrees of freedom. The environment also offers the possibility of many emergent strategies such as using the table to reorient the cube, or using gravity to reorient the cube. Therefore, a large amount of data is required to attain good performance on these tasks. Following \citet{dexPBT} we use the number of successes as a performance metric on these tasks. Note that the DexPBT baseline directly optimizes for success by mutating the reward scales to achieve higher success rate, whereas our method can only optimize a fixed reward function. 
Despite this, we see that \ours achieves a $12-66\%$ higher success rate than DexPBT on regrasping, throw and reorientation. \ours performs $66\%$ better than PBT on the challenging reorientation task. \ours fairs even better on the two-arm reorientation, obtaining more than twice the number of successes on average compared to PBT. Note that vanilla PPO and PQL are unable to learn any useful behaviors on these hard tasks. 

\subsection{In-hand reorientation}

The AllegroHand and ShadowHand reorientation tasks from \citet{li2023parallel} are comparatively easier since they have lower degrees of freedom and the object doesn't move around much and remains inside the hand. On these tasks, we observe that PQL and PPO are able to make significant progress. In particular, we find that PQL is very sample-efficient because it is off-policy and utilizes past data for updates. However, we find that \ours achieves higher \textit{asymptotic performance}. This is because on-policy methods are better at latching onto high reward trajectories and do not have to wait several iterations for the Bellman backup to propagate back to initial states. As discussed previously, in large-scale settings in simulation, we are primarily concerned with asymptotic performance since we want to maximize the downstream performance of our agents (within a reasonable training time budget). We see that on AllegroHand, \ours beats PQL by a 21\% margin, while on the ShadowHand task it achieves comparable performance. On these tasks, both PBT and PPO generally perform worse. This is because PPO is not able to efficiently leverage the large batch size. PBT loses the benefit of its hyperparameter mutation because the environment is simpler and the default hyperparameters work well, so it roughly reduces to simple PPO in $\frac{N}{M}$ environments.  

\begin{figure*}[t!]
    \centering
    \includegraphics[width=\textwidth,height=4cm]{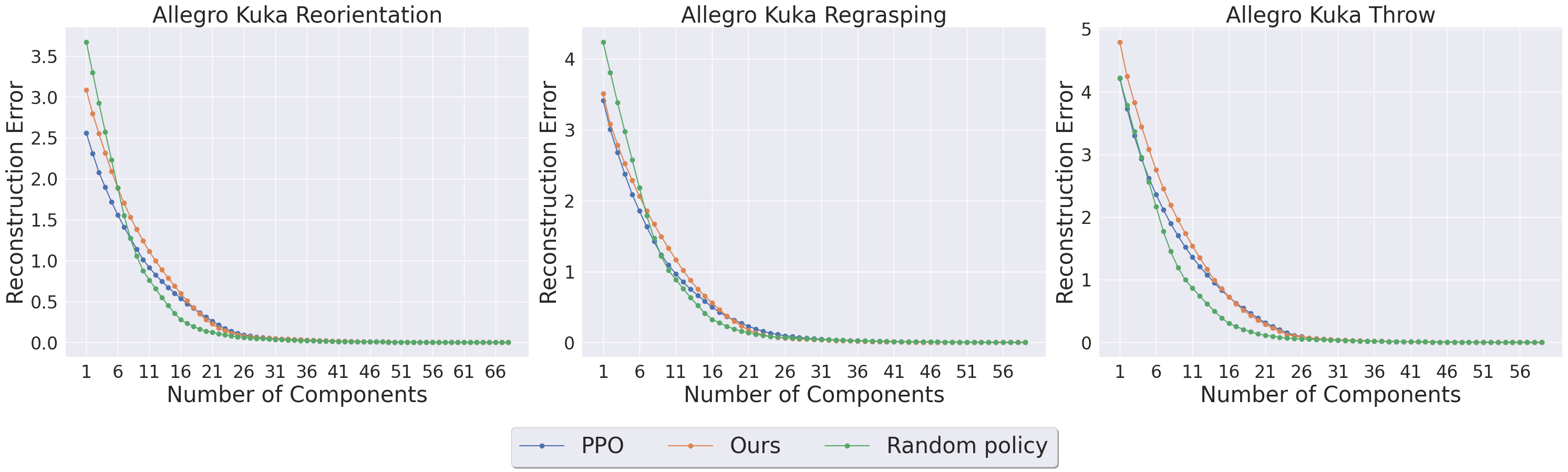}

    \caption{Curves comparing reconstruction error for states visited during training using top-$k$ PCA components for SAPG (Ours), PPO and a randomly initialized policy}
    \label{fig:pca-div}
    % \vspace{-1em}
\end{figure*}

\begin{figure*}[t!]
    \centering
    \includegraphics[width=\textwidth,height=4cm]{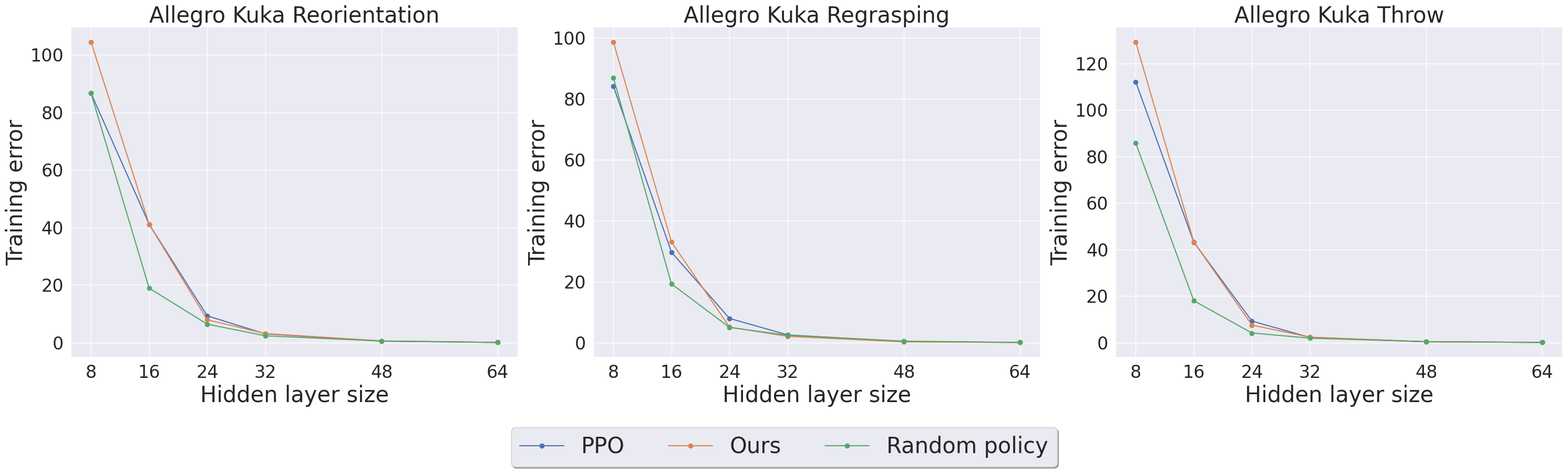}

    \caption{Curves comparing reconstruction error for states visited during training using MLPs with varying hidden layer dimensions for SAPG (Ours), PPO and a randomly initialized policy}
    \label{fig:mlp-div}
    % \vspace{-1em}
\end{figure*}

\subsection{Ablations}
\label{sec:ablations}

The core idea behind \ours is to combine data from different policies instead of optimizing a single policy with an extremely large batch. In this section, we will analyze our specific design choices for how we combine data (choice of $i$ and $\mathcal{X}$ and $\lambda$) and for how we enforce diversity among the data collected by the policies. In particular, we have the following variants
\vspace{-3pt}
\begin{itemize}[noitemsep,left=0pt]
    \item \textbf{\ours (with entropy coef)} As discussed in sec.~\ref{sec:baselines}
 , here we add an entropy loss to the followers to encourage data diversity. We explore different choices for the scaling coefficient of this loss $\sigma\in\{0, 0.005, 0.003\}$.  
 \item \textbf{\ours (high off-policy ratio)} In \ours, when updating the leader, we subsample the off-policy data from the followers such that the off-policy dataset size matches the on-policy data. This is done because off-policy data is typically noisier and we do not want to drown out the gradient from on-policy data. In \ours with a high off-policy ratio, we remove the subsampling step and instead see the impact of computing the gradient on the \textit{entire} combined off-policy + on-policy dataset. 
 \item \textbf{Ours (symmetric)} In \ours, we choose $i=1$ to be the ``leader" and the rest are ``followers". Only the leader receives off-policy data while the followers use the standard on-policy loss. A natural alternative is where there are no privileged policies and each policy is updated with off-policy data from all others as discussed in sec.~\ref{sec:symmetric_agg}. 
 \end{itemize}

 We observe that \ours outperforms or achieves comparable performance to the entropy-regularized variant except in the Reorientation environment where the variant with coefficient $5\mathrm{e-3}$ performs up to $16.5\%$ better. Reorientation is one of the harder tasks out of the four AllegroKuka tasks and has a lot of scope for learning emergent strategies such as using the table to move the object around, etc. Explicit exploration might be useful in discovering these behaviors. 

 The variant of ours which uses all the off-policy data is significantly worse on the AllegroHand and ShadowHand tasks and marginally worse on Regrasping and Throw environments. It is more sample efficient than \ours on Reorientation but achieves lower asymptotic performance. This could be because in the simple environments, additional data has marginal utility. In the harder AllegroKuka environments, it is beneficial to use all the data initially since it may contain optimal trajectories that would otherwise be missed. However, once an appreciable level of performance is achieved, it becomes better to subsample to prevent the noise in the off-policy update from drowning out the on-policy gradient.

 Finally, the symmetric variant of our method performs significantly worse across the board. This is possibly because using all the data to update each policy leads to them converging in behavior. If all the policies start executing the same actions, the benefit of data diversity is lost and \ours reduces to vanilla PPO. Of course, there is a rich space of possible algorithms depending on particular choices of how data is aggregated and diversity is encouraged of which we have explored a small fraction.

\subsection{Diversity in exploration}
To analyze why our method outperforms the baseline method, we conduct experiments comparing the diversity of states visited by each algorithm during training. We devise two metrics to measure the diversity of the state space and find that our method beats PPO in both metrics.

\begin{itemize}
    \item \textbf{PCA} - We compute the reconstruction error of a batch of states using k most significant components of PCA and plot this error as a function of k. In general, a set that has variation along fewer dimensions of space can be compressed with fewer principal vectors and will have lower reconstruction error. This metric therefore measures the extent to which the policy explores different dimensions of state space. Figure-\ref{fig:pca-div} contains the plots for this metric. We find that the rate of decrease in reconstruction error with an increase in components is the slowest for our method. 
    \vspace{-0.35em}
    \item \textbf{MLP} - We train feedforward networks with small hidden layers on the task of input reconstruction on batches of environment states visited by our algorithm and PPO during training. The idea behind this is that if a batch of states has a more diverse data distribution then it should be harder to reconstruct the distribution using small hidden layers because high diversity implies that the distribution is less compressible. Thus, high training error on a batch of states is a strong indicator of diversity in the batch. As can be observed from the plots in Figure-\ref{fig:mlp-div}, we find that training error is consistently higher for our method compared to PPO across different hidden layer sizes.
\end{itemize}
\section{Conclusion}
In this work, we present a method to scale reinforcement learning to utilize large simulation capacity. We show how current algorithms obtain diminishing returns if we perform naive scaling by batch size and do not use the increased volume of data efficiently. Our method achieves state-of-the-art performance on hard simulation benchmarks. 


\section*{Acknowledgements}
We thank Alex Li and Russell Mendonca for fruitful discussions regarding the method and insightful feedback. We would also like to thank Mihir Prabhudesai and Kevin Gmelin for proofreading an earlier draft. This project was supported in part by ONR N00014-22-1-2096 and NSF NRI IIS-2024594.

\bibliography{references}
\bibliographystyle{plainnat}

%%%%%%%%%%%%%%%%%%%%%%%%%%%%%%%%%%%%%%%%%%%%%%%%%%%%%%%%%%%%%%%%%%%%%%%%%%%%%%%
%%%%%%%%%%%%%%%%%%%%%%%%%%%%%%%%%%%%%%%%%%%%%%%%%%%%%%%%%%%%%%%%%%%%%%%%%%%%%%%
% APPENDIX
%%%%%%%%%%%%%%%%%%%%%%%%%%%%%%%%%%%%%%%%%%%%%%%%%%%%%%%%%%%%%%%%%%%%%%%%%%%%%%%
%%%%%%%%%%%%%%%%%%%%%%%%%%%%%%%%%%%%%%%%%%%%%%%%%%%%%%%%%%%%%%%%%%%%%%%%%%%%%%%
\newpage
\appendix
\onecolumn
\section{Task and Environment Details}
\label{appendix:env}

\paragraph{Hard Difficulty Tasks}
All four hard tasks are based on the \texttt{Allegro-Kuka} environments\citep{dexPBT}. These consist of an Allegro Hand (16 dof) mounted on a Kuka arm (7 dof). In each case, the robot must manipulate a cuboidal kept on a fixed table. The observation space is $\mathbf{o}_t = [\mathbf{q}, \dot{\mathbf{q}}, \mathbf{x}_t, \mathbf{v}_t, \mathbf{\omega}_t, \mathbf{g}_t, \mathbf{z}_t]$, where $\mathbf{q}$, $\dot{\mathbf{q}} \in \mathbb{R}^{23}$ are the joint angles and velocities respectively of each joint of the robot, $\mathbf{x}_t \in \mathbb{R}^7$ is the pose of the object, $\mathbf{v}_t$ is its linear velocity and $\mathbf{\omega}_t$ is its angular velocity, $\mathbf{g}_t$ is a task-dependent goal observation and $\mathbf{z}_t$ is auxiliary information pertinent to solving the task such as if the object has been lifted. These tasks consist of a complex environment but a simple reward function allowing opportunities for emergent strategies to be learnt such as in-hand reorientation under the influence of gravity, reorientation against the table, different types of throws and grasps and so on. The performance of the agent in the above three tasks is measured by the successes metric which is defined as the number of successes in a single episode. Three tasks include:

\begin{itemize}
    \item \textbf{Regrasping} - The object must be lifted from the table and held near a goal position $\mathbf{g}_t \in \mathbb{R}^3$ for $K=30$ steps. This is called a ``success". The target position and object position are reset to a random location after every success. The success tolerance $\delta$ defines the maximum error between object pose and goal pose for a success $\|\mathbf{g}_t - (\mathbf{x}_t)_{0:3}\| \leq \delta$. This tolerance is decreased in a curriculum from 7.5cm to 1cm, decremented by 10\% each time the the average number of successes in an episode crosses 3. The reward function is a weighted combination of rewards encouraging the hand to reach the object $r_{reach}$ , a bonus $r_{lift}$, rewards encouraging the hand to move to goal location after lifting $r_{target}$ and a success bonus $r_{success}$. 
    \item \textbf{Throw} - The object must be lifted from the table and thrown into a bucket at $\mathbf{g}_t\in\mathbb{R}^3$ placed out of reach of the arm. The bucket and the object position are reset randomly after every successful attempt. The reward function is similar to regrasping with the difference being that the target is now a bucket instead of a point.
    \item \textbf{Reorientation} - This task involves picking up the object and reorienting it to a particular target pose $\mathbf{g_t}\in\mathbb{R}^7$ (position + orientation). Similar to the regrasping task, there is a success tolerance $\delta$ which is varied in a curriculum. The target pose is reset once the agent succeeds. This means the agents needs to the object in different poses in succession, which may sometimes entail placing the obxfject on the table and lifting it up in a different way. Here too, the reward function is similar to regrasping, with the goal now being a pose in $\mathbb{R}^7$  instead of $\mathbb{R}^3$.

    \item \textbf{Two Arms Reorientation}: The objective is the same to the AllegroKuka Reorientation task. The difference is that the setup has another arm now. In this task, the target pose may be within the reach of one arm but not the other, which means that the arms may need to transfer the object between themselves, for which it needs to learn complex throwing and catching behaviours.
\end{itemize}

\paragraph{Easy Difficulty Tasks:}
In addition, we test on the following dexterous hand tasks. As before, the observation space consists of the joint angles and velocities $\mathbf{q}_t$, $\dot{\mathbf{q}}_t$, object pose $\mathbf{x}_t$ and velocities $\mathbf{v}_t, \mathbf{\omega}_t$. Following previous works \citep{li2023parallel}, we use the net episode reward as a performance metric for the ShadowHand and AllegroHand tasks.

\begin{itemize}
    \item \textbf{Shadow Hand}: We test on in-hand reorientation task of a cube using the 24-DoF Shadow Hand(\citep{openai_shadow_hand}). The task is to attain a specified goal orientation (specified as a quaternion) for the cube $\mathbf{g}_t \in \mathbb{R}^4$. The reward is a combination of the orientation error and a success bonus. 
    \item \textbf{Allegro Hand}: This is the same as the previous in-hand reorientation task but with the 16-DoF Allegro Hand instead. 
\end{itemize}

\section{Training hyperparameters}
%%%%%%%%%%%%%%%%%%%%%%%%%%%%%%%%%%%%%%%%%%%%%%%%%%%%%%%%%%%%%%%%%%%%%%%%%%%%%%%
%%%%%%%%%%%%%%%%%%%%%%%%%%%%%%%%%%%%%%%%%%%%%%%%%%%%%%%%%%%%%%%%%%%%%%%%%%%%%%%
We use two different sets of default hyperparaeters for PPO in AllegroKuka and Shadow Hand tasks which are descibed below.

\subsection{AllegroKuka tasks}
We use a Gaussian policy where the mean network is an LSTM with 1 layer containing 768 hidden units. The observation is also passed through an MLP of with hidden layer dimensions $768 \times 512 \times 256$ and an ELU activation \citep{elu} before being input to the LSTM. The sigma for the Gaussian is a fixed learnable vector independent of input observation.

\begin{table}[H]
    \centering
    \begin{tabular}{c|c}
    \hline
    \textbf{Hyperparameter} & \textbf{Value} \\
    \hline
       Discount factor, $\gamma$  & 0.99  \\
       $\tau$  & 0.95 \\
       Learning rate & 1e-4 \\
       KL threshold for LR update & 0.016 \\
       Grad norm & 1.0 \\
       Entropy coefficient & 0 \\ 
       Clipping factor $\epsilon$ & 0.1 \\
       Mini-batch size & $\textrm{num\_envs}\cdot 4$ \\ 
       Critic coefficient $\lambda'$ & 4.0 \\
       Horizon length & 16 \\
       LSTM Sequence length & 16 \\
       Bounds loss coefficient & 0.0001 \\
       Mini epochs & 2 \\
    \hline
    \end{tabular}
    \caption{Training hyperparameters for AllegroKuka tasks}
    \label{tab:train_params_kuka}
\end{table}

\subsection{Shadow Hand}
We use a Gaussian policy where the mean network is an MLP with hidden layers dimensions $512\times 512 \times 256 \times 128$ and an ELU activation \citep{elu}  

\begin{table}[H]
    \centering
    \begin{tabular}{c|c}
    \hline
    \textbf{Hyperparameter} & \textbf{Value} \\
    \hline
       Discount factor, $\gamma$  & 0.99  \\
       $\tau$  & 0.95 \\
       Learning rate & 5e-4 \\
       KL threshold for LR update & 0.016 \\
       Grad norm & 1.0 \\
       Entropy coefficient & 0 \\ 
       Clipping factor $\epsilon$ & 0.1 \\
       Mini-batch size & $\textrm{num\_envs}\cdot 4$ \\ 
       Critic coefficient $\lambda'$ & 4.0 \\
       Horizon length & 8 \\
       Bounds loss coefficient & 0.0001 \\
       Mini epochs & 5 \\
    \hline
    \end{tabular}
    \caption{Training hyperparameters for Shadow Hand}
    \label{tab:train_params_shadow}
\end{table}

\subsection{Allegro Hand}
We use a Gaussian policy where the mean network is an MLP with hidden layers dimensions $512 \times 256 \times 128$ and an ELU activation.  

\begin{table}[H]
    \centering
    \begin{tabular}{c|c}
    \hline
    \textbf{Hyperparameter} & \textbf{Value} \\
    \hline
       Discount factor, $\gamma$  & 0.99  \\
       $\tau$  & 0.95 \\
       Learning rate & 5e-4 \\
       KL threshold for LR update & 0.016 \\
       Grad norm & 1.0 \\
       Entropy coefficient & 0 \\ 
       Clipping factor $\epsilon$ & 0.2 \\
       Mini-batch size & $\textrm{num\_envs}\cdot 4$ \\ 
       Critic coefficient $\lambda'$ & 4.0 \\
       Horizon length & 8 \\
       Bounds loss coefficient & 0.0001 \\
       Mini epochs & 5 \\
    \hline
    \end{tabular}
    \caption{Training hyperparameters for Shadow Hand}
    \label{tab:train_params_allegro}
\end{table}

\paragraph{Note:} In case of experiments with entropy based exploration, each block of environments has it's own learnable vector sigma which enable policies for different blocks to have different entropies.
\end{document}